\documentclass[11pt]{article}

\usepackage[preprint]{acl}

\usepackage{times}
\usepackage{latexsym}

\usepackage[T1]{fontenc}

\usepackage[utf8]{inputenc}

\usepackage{microtype}

\usepackage{inconsolata}

\usepackage{graphicx}

\usepackage{booktabs}

\DeclareCaptionLabelFormat{appendixformat}{Appendix #1 #2}

\title{LLMs as survey respondents: Comparison to simple semantic baselines}

\author{Esko Lehtonen \\
  VTT Technical Research Centre of Finland Ltd \\ Espoo, Finland  \\
  \texttt{esko.lehtonen@vtt.fi} \\\And
  Tommi Buder-Gr\"{o}ndahl \\
  University of Helsinki \\ Helsinki, Finland \\
  \texttt{tommi.grondahl@helsinki.fi} \\}

\begin{document}
\maketitle
\begin{abstract}
Large language models (LLMs) have been explored as surrogates for human survey respondents. A more promising application is data augmentation, where LLMs predict new responses from partial human data. However, the semantic similarity between questionnaire items is a known source of correlation between answers. It is thus unclear whether LLMs actually infer meaningful relationships between question-response pairs, or use surface-level comparison between questions as a heuristic. To assess this, we evaluated how three LLMs can predict responses to held-out items in two surveys measuring attitudes toward automated or light-electric vehicles, based on varying proportions of known responses. Performance was compared to two baselines, where new responses were derived by maximizing embedding-similarity to a prior response. While LLMs mostly predicted human responses better than baselines, their performance consistently declined with a lower share of known responses, even aligning with the baseline in one survey. Our results show that LLMs' capacities for augmenting replies varied across surveys, and can revert to the same levels as simpler embedding similarity baselines when less data is available.

\end{abstract}

\section{Introduction}

Given the costs of conducting surveys, the prospect of using LLM surrogates has gained traction. These can mimic humans to a notable degree, especially when conditioned with contextual information \citep{Bisbeeetal24, Dashetal25, Zhaoetal25, Zhangetal26}. Potential use cases involve generating completely new responses or augmenting existing data with novel but related questions \cite{wallius_beyond_2026}. 

However, a foundational matter is yet to be investigated in this context: \emph{why} do LLMs mimic human responses? Do they perform inferences using internalized knowledge, or base their answers on something more superficial?
In particular, it has been proposed that survey responses are predictable from the semantic similarity between questions \citep{Arnulfetal14, Arnulfetal24}.
If so, the performance of LLM surrogates might simply reflect this rather than a genuine understanding of the survey topic.

To address this, we compared survey responses given by three LLMs and two baselines relying on sentence embedding similarities \citep{Lehtonenetal25}.
We applied each model to two surveys measuring attitudes toward automated vehicles: AV \citep{Nordhoffetal21} and LEV \citep{mesimaki_light_2023}. Each experiment was run in three variants, predicting novel responses from approximately $20\%$, $50\%$, or $80\%$ of known human responses.

LLMs generally exceeded both baselines in alignment with human responses and sensitivity to the known response rate. However, the performance of all models was systematically lower for LEV, and the baselines even matched LLMs with the known response rate of 20\%. We explore possible reasons for this, and their ramifications for using LLMs for survey data augmentation going forward.

\section{Background}

\subsection{LLMs as survey respondents}

As benefits of using LLMs for generating survey responses,  \citet{Jansenetal23} highlight speed, price, scalability, and the relative consistency across replies \citep{Gilardietal23}.
Concurrently, prominent concerns include biases \citep{Kitadaietal24}, model opacity \citep{Eslamietal19, Ollionetal24}, 
reproducibility \citep{Bisbeeetal24}, struggles with rare domain-specific terms \citep{Morozovskii:Ramanna23}, and partial deviations from human responses.

\citet{Bisbeeetal24} compared several ChatGPT 3.5 Turbo ``personas'' to corresponding sociopolitical groups. Synthetic data exhibited less variation and several statistically significant differences from human responses. 
\citet{Dominguez-Olmedoetal24} evaluated 43 LLMs on questions from the American Community Survey, and found ordering biases for the prompt as well as systematically higher entropy than in human responses. 
\citet{Kitadaietal24} showed discrepancies between humans and GPT-4o-mini in responding to a Japanese financial questionnaire, taking these to illustrate cultural biases inherent in LLMs \citep{Liangetal21}.

Nevertheless, additional fine-tuning or prompt engineering can increase human-LLM alignment.
\citet{Caoetal25} compared the performance of several models in the World Values Survey \citep{Haerpferetal22},  where fine-tuning by first-token alignment with human replies incurred a major improvement over zero-shot performance.
\citet{Dashetal25} used Low-Rank Adaptation \citep{Huetal21} to tune LLMs for responding based on PersonHub profiles \citep{Geetal24},
which allowed smaller and larger models to reach similar levels of consistency.
\citet{Zhaoetal25} found the enhancement of prompts with contextual information to significantly improve the response consistency of four LLMs (GPT-3.5 Turbo, GPT-4 Turbo, LLaMA 3.0-8B, LLaMA 3.1-8B).
\citet{Zhangetal26} improved human-LLM alignment for three models (GPT 4.1, Gemini 2.0 Flash, Claude 3.7 Sonnet) with persona prompting.
However, \citet{Morochoetal26} found the effects of persona prompting to be limited and partly even negative due to distortions.

LLMs have been explored for missing value imputation \cite{mangussi_large_2026, kim_ai-augmented_2026, ji_predicting_2024}. Instead of replacing human participants, response-conditioned LLMs could augment dataset by inferring novel answers from existing ones. 

\subsection{Semantic predictability of responses}

\citet{Arnulfetal14} put forward the \emph{semantic theory of survey response} (STSR), which states that human answers to surveys can be predicted to a significant degree from the semantic similarity between questions. Using Latent Semantic Analysis (LSA) \citep{Foltz96} and a corpus-based text similarity algorithm \citep{Mihalceaetal26}, \citeauthor{Arnulfetal14} could explain $60\%-86\%$ of variance in human answers to the Multifactor Leadership Questionnaire \citep{Avolio:Bass04}.
\citet{Gefen:Larsen17} further partly replicated the Technology Acceptance Model \citep{Davis89} using LSA. Surveying prior work (e.g. \citealt{Kjelletal19, Rosenbusch20, Larsenetal23}), \citet{Arnulfetal24} contend that semantic similarity could predict human response patterns with $80\%-90\%$ accuracy.

Recently, \citet{Lehtonenetal25} used Sentence-BERT \citep{Reimers:Gurevych19} to generate synthetic answers to a questionnaire concerning automated vehicles \citep{Nordhoffetal21}. Based on selecting the most similar response to a prior one in embedding distance, the synthetic data accounted for $40\%$ of between variable correlations (covariance) in human answers.

\section{Methods}

\subsection{Data}

We tested LLMs' ability to perform response-conditioned imputation against human data from two surveys. \textit{AV} investigates the acceptance of automated driving \cite{Nordhoffetal21}. A subset of 22 questions was used, measuring six factors \cite{Lehtonenetal25}.\footnote{Behavioural Intention, Facilitation Conditions, Hedonic Motivation, Performance Expectancy, Social Influence, Trust.} A sample of 500 respondents were drawn at random, derived from the data in Zenodo \cite{nordhoff_2023_8389718}. \textit{LEV} investigates the benefits of light-electric vehicles \cite{mesimaki_light_2023}.
The original 646 respondents were pruned to 426 by removing those where the response was missing or ``don't know''.
A subset of 21 questions was used, forming four factors\footnote{Accessibility, Safe, Comfort, Practicality.} along with eight additional questions.
Unlike AV, LEV is not publicly available online.

Three masked datasets were created for both surveys by varying the share of known responses (20\%, 50\%, 80\%, rounded to the nearest integer). Question order was randomized for each participant, and answers were masked from the end. The six masked datasets were used for generating responses, providing identical inputs between models.  

\subsection{Response generation}

Three LLMs (Gemini 3.7 Flash, GPT 5.6 Luna, Mistral 4 Small) were used with a system prompt to generate consistent answers based on known question-answer pairs (see Appendix~\ref{app:llm_robustness}). The task prompt instructed the model to provide answers to questions when these were masked, based on known question-answer pairs (see Appendix~\ref{app:task_prompt_example}).\footnote{Mistral generated nine answers with duplicated question IDs, which were removed. No other output errors appeared.}

The semantically derived baselines exemplify the \mbox{\textit{Semagent}} method, where the response is chosen by maximizing its embedding similarity to a prior response \cite{Lehtonenetal25}.
Responses are sentences that combine the question and answer: \mbox{\textit{I <statement> that <answer>}}.
The algorithm was initialized with ``memorized'' responses from known question-answer pairs.\footnote{This contrasts the original method of \citet{Lehtonenetal25}, where memorized responses were selected at random. We also did not add noise to the responses.}
An embedding similarity matrix across possible responses was created for both SBERT \citep{Reimers:Gurevych19}
and Gemini embeddings.
The response was greedily selected based on the most similar prior response, as other means did not consistently improve performance (see Appendix \ref{app:Semagent_variants}.

\subsection{Analysis}

Alignment between human and model-generated data was estimated with Cohen's kappa ($\kappa$) \cite{bendavid_comparison_2008}, calculated only from predicted answers.
Squared weighting was used since responses were ordinal. 95\% confidence intervals (CIs) were 
bootstrapped with 500 iterations and bias-corrected and accelerated (BCa) bootstrap intervals. Respondent-level clustering was used since multiple responses were from each participant violate the assumption of independent observations.
Statistical inferences were performed by bootstrapping the difference in $\kappa$ statistics ($\Delta\kappa$). Bonferroni correction was applied with multiple comparisons (See Appendix \ref{app:pairwise_tests} for further details).

Covariance recovery was assessed with covariance matrices. Frobenius norms for the difference between human and imputed datasets measure their Euclidean distance. RV coefficients between these matrices measure the similarity of their shapes. Factor structure recovery was investigated with confirmatory factor analysis (CFA). Previously validated factor structures were used for both human and imputed datasets. 

\section{Results}

\subsection{Alignment with human data}

\begin{figure*}[h]
  \centering
  \includegraphics[width=1.0\linewidth]{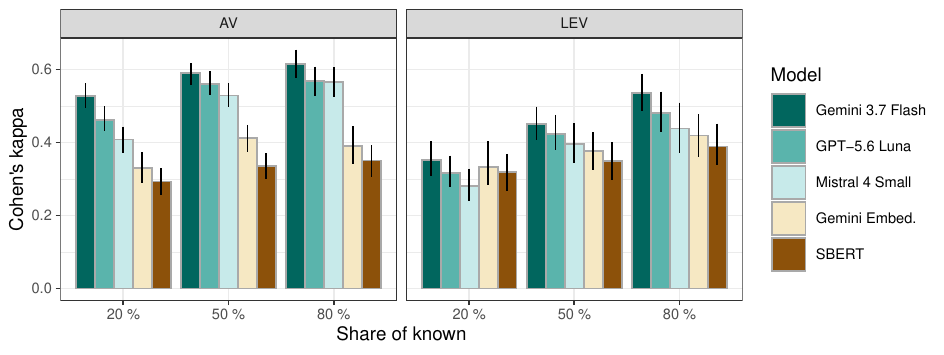}
  \caption{$\kappa$ between human and generated responses by different shares of known responses. Means and $95\%$ CIs.}
  \label{fig:split_kappa_predicted}
\end{figure*}

Figure \ref{fig:split_kappa_predicted} shows $\kappa$ for each model with different shares of known answers. $\Delta\kappa$ values with 95\% CIs for the pairwise contrasts are reported in Appendix~\ref{app:pairwise_tests}.

LLMs show a systematic pattern with 50\% known response rate: Gemini 3.7 Flash outperforms GPT 5.6 Luna, and GPT 5.6 Luna outperforms Mistral 4 Small. With 20\% and 80\% known response rates, the pattern was repeated but not all pairwise differences were statistically significant. 
With Semagent baselines, Gemini produced more aligned responses than SBERT for AV with 50 \% and 80 \% of known answers. Otherwise the difference was not statistically significant. 
LLMs always outperformed the best Semagent (Gemini) in AV. For LEV, the LLM advantage was more limited: Gemini 3.7 Flash outperformed baselines with 50\% and 80\% known response rates, and GPT 5.6 Luna with 50\%.

All LLMs were more aligned in AV compared to LEV with 20\% and 50\% of known responses, but only Mistral 4 Small with 80\%. Semagents' performance was not different between datasets. 

\subsection{Known response rates}

LLMs produced more aligned responses with a known response rate of 50\%  rather than 20\%, across both datasets. The increase from 50\% to 80\% also improved alignment, but this was statistically significant only for Gemini 3.7 Flash in LEV. An improvement from 20\% to 50\% was found in Semagent baselines as well, but only in AV.

\subsection{Covariance and factor structure}

Increasing known responses improved covariance preservation both in terms of the shape and magnitude (Table~\ref{table:rv_frobenius}) of the covariance matrix. With a 20\% rate, LLMs had an advantage over Semagents in L3Pilot. Otherwise, LLMs did not show a strong improvement over Semagents. 

\begin{table*}[t]
\caption{RV coefficient(range: $0-1$, higher is better) and Frobenius norms (lower is better) for imputed datasets.} 
\label{table:rv_frobenius}
\small
\begin{tabular*}{\linewidth}{@{\extracolsep{\fill}}lrrrrrrrrrrrr}
\toprule
 & \multicolumn{6}{c}{{RV coefficient}} & \multicolumn{6}{c}{{Frobenius norm}} \\
\cmidrule(lr){2-7} \cmidrule(lr){8-13}
 & \multicolumn{3}{c}{{AV}} & \multicolumn{3}{c}{{LEV}} & \multicolumn{3}{c}{{AV}} & \multicolumn{3}{c}{{LEV}} \\
\cmidrule(lr){2-4} \cmidrule(lr){5-7} \cmidrule(lr){8-10} \cmidrule(lr){11-13}
Model & 20 \% & 50 \% & 80 \% & 20 \% & 50 \% & 80 \% & 20 \% & 50 \% & 80 \% & 20 \% & 50 \% & 80 \% \\
\midrule\addlinespace[2.5pt]
Gemini 3.7 Flash & 0.94 & 0.98 & 1.00 & 0.91 & 0.97 & 1.00 & 5.50 & 3.16 & 1.28 & 5.68 & 2.82 & 1.08 \\
GPT-5.6 luna & 0.91 & 0.97 & 1.00 & 0.87 & 0.97 & 0.99 & 5.86 & 3.58 & 1.41 & 6.63 & 3.31 & 1.39 \\
Mistral 4 Small & 0.93 & 0.97 & 0.99 & 0.89 & 0.98 & 1.00 & 5.31 & 3.94 & 1.79 & 5.39 & 2.49 & 0.84 \\
Gemini Embed. & 0.74 & 0.94 & 0.99 & 0.95 & 0.98 & 0.99 & 11.71 & 5.14 & 2.29 & 5.00 & 2.49 & 1.30 \\
SBERT & 0.71 & 0.90 & 0.98 & 0.91 & 0.96 & 0.99 & 14.42 & 6.38 & 2.74 & 11.26 & 4.26 & 1.51 \\
\bottomrule

\end{tabular*}
\end{table*}

LLM-imputed datasets contained the same factor structure as human data with 50\%--80\% of known answers (see Appendix ~\ref{app:factor_structure}). With 20\%, only Mistral 4 Small converged. \mbox{Semagents} never converged. Generated data shows systematic item-level biases. 

\section{Discussion}

Excluding LEV with $20\%$ known response rate, all LLMs outperformed both Semagents across all experiment variants. Of the Semagents, Gemini Embeddings systematically performed better than SBERT. As expected, a higher share of known responses increased alignment with human responses as well as covariance preservation. However, both Semagents were impacted less by known response rate than LLMs. For LLMs, the increase from $20\%$ to $50\%$ had a larger impact than the increase from $50\%$ to $80\%$. With $50\%-80\%$ known response rates, LLM and human data aligned in factor structure.

The two surveys diverge in how LLMs relate to Semagents. All LLMs had a higher alignment with human data in AV, and also LLMs' advantage was greater. LLMs in AV markedly increased from $20\%$ to $50\%$, but remained relatively stagnant after this.  In LEV, LLM's performance was closer to Semagents, but increased more steadily with known response rates. Finally, LEV with $20\%$ known response rate constitutes a crucial outlier among our results: all models had poor performance, and the otherwise clear distinction between LLMs and Semagents is absent.

The results may be influenced by AV being open access while LEV is not. Open datasets can be present in the training data of LLMs, artificially facilitating their performance \cite{mangussi_large_2026}. AV might also be similar to other  datasets that LLMs have been exposed to during training, since its topic -- automated vehicle acceptance -- is widely studied. LEV, in turn, concerns a more specialized theme on with fewer studies. Hence, with LEV, overlap with training data is impossible and model generalization based on similar training data is less likely.
With 20\% of known answers, LLM performance in LEV closely matched that of Semagents. In line with the STSR, this may indicate that LLMs were unable to use memorized question-answer pairs and reverted to using generic semantic information as Semagents.

Together, the impact of known answer rate and the contrast between datasets suggest that LLMs' access to both contextual and memorized information is key for explaining their performance.

\section{Conclusions}

We draw two key takeaways from our results.
First, the response augmentation capacities of LLMs can vary significantly across surveys and known response rates. Second, comparison to embedding similarity baselines indicates that LLMs do not always exceed the performance of simpler semantic similarity metrics. This is in line with the STSR, which states that survey responses are substantially predictable from the semantics of questions. Assessing it has been rare in prior work on LLM surrogates, and we stress the need to include it as part of the experimental setup going forward. 

\newpage

\section*{Limitations}

This short paper has experimented on three English-dominant LLMs (Gemini 3.7 Flash, GPT 5.6 Luna, and Mistral 4 Small) and two \mbox{Semagent} baselines (built on SBERT all-mpnet-v2 and Gemini embeddings). The survey questions were formulated in English and concerns a highly specific domains. For broader conclusions to be drawn, a range of similar evaluations are needed at a broader scale, encompassing a variety of survey domains, model architectures, fine-tuning and prompting regimes, as well as different languages and sociodemographic groups. 

\section*{Ethical considerations}

All research procedures and protocols are exempt from review board approval according to the applicable guidelines. Publicly available survey data contained no personal data. Respondents were adults, who gave informed consent for participation and use of their responses for research. The results indicate that using LLMs or semantic baselines for data augmentation may introduce systematic item-level biases, which should be considered.

\section*{Code and data availability}
Analysis code will be made public, if the paper is accepted. AV dataset is openly available at Zenodo \cite{nordhoff_2023_8389718}. Inquiries regarding the LEV dataset to the first author.    

\section*{Information about use of AI assistants}
The first author used Google Gemini and Microsoft Copilot CLI to assist in modifying and debugging the software code. The author reviewed and edited the content as needed and takes full responsibility for the content of the published article.

\bibliography{refs}

\clearpage
\appendix

\setcounter{figure}{0}
\setcounter{table}{0}
\renewcommand{\thefigure}{\arabic{figure}}
\renewcommand{\thetable}{\arabic{table}}
\captionsetup[figure]{labelformat=appendixformat}
\captionsetup[table]{labelformat=appendixformat}

\onecolumn
\section{System prompt and its robustness}
\label{app:llm_robustness}

SP-0 system prompt was used for generating the answers. Three variants of the system prompt instructing the LLMs to generate consistent answers were tested to check if answer generation was robust against small variation in the instruction (SP-1--SP-4). The variant SP-4 asking for inconsistent responses was used to demonstrate that the system prompts did influence the response behaviour of LLMs:

\begin{description}

    \item[\textbf{SP-0}] You are a consistent survey respondent. Your goal is to answer new questions based on what can be inferred from the answers to the previous ones.

    \item[\textbf{SP-1}] You are a consistent survey respondent. Your goal is to answer new questions while maintaining a coherent personality, logic, and set of values based on your previous responses.
    
    \item[\textbf{SP-2}] You are a consistent survey respondent. Your goal is to answer new questions based on what follows from the answers to the previous ones.
    
    \item[\textbf{SP-3}] You are a consistent survey respondent. Your goal is to provide responses that are logically consistent and aligned with the values, lifestyle, and opinions established in your previous answers.
    
    \item[\textbf{SP-4}] You are an inconsistent survey respondent. Your goal is to answer new questions contradicting your previous answers.
\end{description}

System prompt variants and models were compared using data with 50\% of known answers with 100 participants. The differences between consistent system prompt variants were small (Appendix Table~\ref{table:kappa_LLM_variants}). With inconsistent prompt (SP-4) , the negative $\kappa$ values showed clear disagreement between LLM and human answers. This shows that LLMs were robust for small difference in the consistent prompts, and that the system prompt had an effect on the answers. 

Temperature and reasoning effort hyperparameters were explicitly set for LLMs (Appendix Table~\ref{table:llm_hyperparameters}). 

\begin{table*}[t]
\caption{$\kappa$ and 95\% CI for LLM generated data by varying the system prompt with 100 respondents.}
\label{table:kappa_LLM_variants}
\small
\begin{tabular*}{\linewidth}{@{\extracolsep{\fill}}llrrr}
\toprule
Model & Variant & Kappa & Lower CI & Upper CI \\
\midrule\addlinespace[2.5pt]
GPT-5.6 Luna & SP-0 & 0.570 & 0.506 & 0.640 \\
GPT-5.6 Luna & SP-1 & 0.568 & 0.501 & 0.630 \\
GPT-5.6 Luna & SP-2 & 0.585 & 0.522 & 0.649 \\
GPT-5.6 Luna & SP-3 & 0.577 & 0.514 & 0.641 \\
GPT-5.6 Luna & SP-4 & -0.418 & -0.482 & -0.349 \\
Gemini 3.7 Flash & SP-0 & 0.621 & 0.556 & 0.690 \\
Gemini 3.7 Flash & SP-1 & 0.612 & 0.542 & 0.674 \\
Gemini 3.7 Flash & SP-2 & 0.619 & 0.550 & 0.690 \\
Gemini 3.7 Flash & SP-3 & 0.605 & 0.523 & 0.670 \\
Gemini 3.7 Flash & SP-4 & -0.449 & -0.518 & -0.391 \\
Mistral 4 Small & SP-0 & 0.540 & 0.452 & 0.609 \\
Mistral 4 Small & SP-1 & 0.564 & 0.500 & 0.633 \\
Mistral 4 Small & SP-2 & 0.532 & 0.457 & 0.599 \\
Mistral 4 Small & SP-3 & 0.553 & 0.488 & 0.632 \\
Mistral 4 Small & SP-4 & -0.327 & -0.410 & -0.268 \\
\bottomrule

\end{tabular*}
\end{table*}

\begin{table} 
    \centering 
    \caption{LLM hyperparameters} 
    \label{table:llm_hyperparameters}
    \begin{tabular}{ll} 
    \toprule Model & Parameters \\ 
    \midrule 
    Gemini 3.7 Flash & \texttt{thinking\_level = medium} \\
    & \texttt{temperature = 1.0} \\ 
    GPT 5.6 Luna & \texttt{reasoning\_effort = medium} \\
    & \texttt{temperature = 1.0} \\ 
    Mistral 4 Small & \texttt{reasoning\_effort = none} \\
    & \texttt{temperature = 0.7} \\
    \bottomrule \end{tabular} 
\end{table}

\clearpage
\onecolumn
\section{Example task prompt}
\label{app:task_prompt_example}

\begin{quote}

\#Historical Responses (Format: Question -> Answer):

I would be able to get help from my friends and/or family when I have difficulties using a conditionally automated car. -> Disagree

Using a conditionally automated car would help me to reach my destination faster. -> Agree

I intend to use a conditionally automated car in the future. -> Agree

The brand of the conditionally automated car would be the most important thing I would consider before purchasing one. -> Disagree

I could acquire the necessary knowledge to use a conditionally automated car. -> Agree

The cost of the conditionally automated car would be the most important thing I would consider before purchasing one. -> Strongly agree

I plan to use a conditionally automated car once it becomes available. -> Agree

I would engage in other tasks while the conditionally automated car is in control. -> Agree

I assume that people whose opinions I value would prefer that I use a conditionally automated car. -> Strongly disagree

Using a conditionally automated car would help me reach my destination more comfortably. -> Neither agree nor disagree

I would not be able to get help from car dealers when I have difficulties using a conditionally automated car. -> Disagree

Using a conditionally automated car would be enjoyable. -> Agree

The benefits of using a conditionally automated car would be the most important thing I would consider before purchasing one. -> Agree

The next car I buy will be a conditionally automated car, if it is available. -> Agree

I find it important that the conditionally automated car has a sleek and cool design. -> Agree

Using a conditionally automated car would help me reach my destination more safely. -> Agree

I would be able to read the driver manual when I have difficulties using a conditionally automated car. -> Agree

It would make me proud to own a conditionally automated car.  -> Neither agree nor disagree 

\#Tasks 

Answer to the following new questions: (Format: ID: Question): 

q12.2: I would trust conditionally automated cars.

q12.4: I would feel hesitant about using a conditionally automated car.

q14r.12: Using a conditionally automated car would give me status and prestige among people important to me.

q12.1: I would be suspicious of conditionally automated cars.

\#Constraint
Use only this scale to respond: Strongly disagree, Disagree, Neither agree nor disagree, Agree, Strongly agree.

\#Output Format
ID: Response (Example: q12.1: Agree)
\end{quote}

\clearpage
\onecolumn
\section{Pairwise tests}
\label{app:pairwise_tests}

Respondent-clustered bootstrapped was used because each participant contributed multiple responses. Respondent-clustering guarantees that all the responses from the same participants are present in the bootstrapped sample. 

For statistical inference, respondent-clustered bootstrapped CIs for the difference in $\kappa$ values were calculated ($\Delta\kappa)$. At each bootstrap sample, $\kappa$ was calculated for both batches. $\Delta\kappa)$ was calculated and used for constructing bias-corrected and accelerated (BCa) bootstrap intervals. $\alpha$ level was set to $0.05$. Bonferroni corrections were applied when multiple comparisons were made (no comparisons --> 95\% CI, two comparisons --> 97.5\% CI, three comparisons --> 98.3\% CI, ...).

For the within-dataset contrasts, the respondent-clustered bootstrapping was performed by selecting the same respondent from the the compared batches to respect the paired nature of the comparison. For between-dataset comparison, the respondents in the two batches are not the same, so the bootstrapping is performed independently. 

\begin{figure}
    \centering
    \includegraphics[width=1\textwidth]{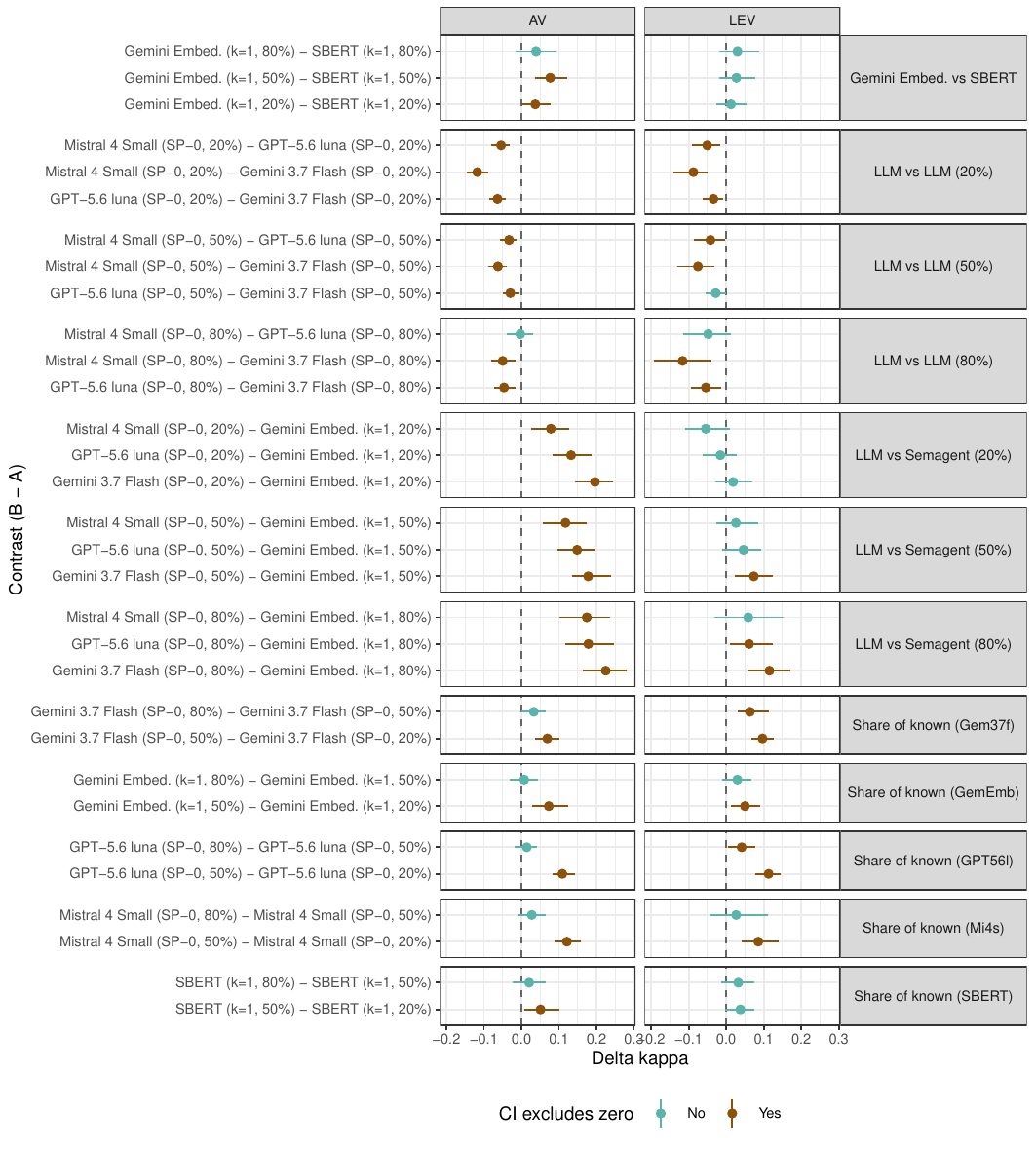}
    \caption{Bootstrapped $\Delta\kappa$ with Bonferroni corrected CIs ($\alpha=0.05$) for within-dataset contrasts. If CI overlap zero the difference is not statistically significant. If CI are above zero, then B had statistically significantly higher $\kappa$ values than A, and vice versa.}
    \label{fig:delta_kappas_within}
\end{figure}

\begin{figure}
    \centering
    \includegraphics[width=0.7\textwidth]{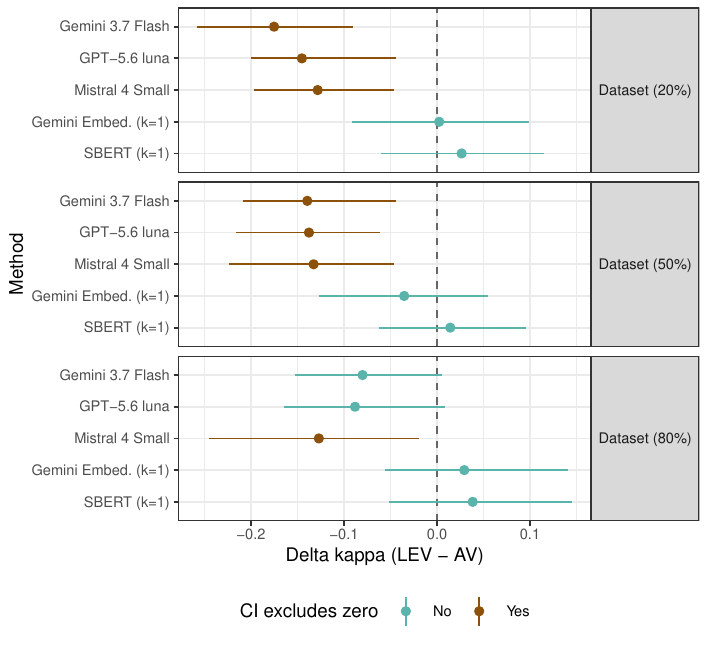}
    \caption{Bootstrapped $\Delta\kappa$ with Bonferroni corrected CIs ($\alpha=0.05$) for between-dataset contrasts. If CI overlap zero the difference is not statistically significant. If CI are above zero, then B had statistically significantly higher $\kappa$ values than A, and vice versa.}
    \label{fig:delta_kappas_between}
\end{figure}

Appendix Figure \ref{fig:delta_kappas_within} shows the $\Delta\kappa$ values with the Bonferroni corrected CIs for the within-dataset contrasts, and Appendix Figure \ref{fig:delta_kappas_between} for the between-dataset contrasts. Bonferroni correction was performed for as many comparisons as shown in each panel. 

\section{Testing Semagent with different neighbourhood sizes}
\label{app:Semagent_variants}

Semagent algorithm can be criticized for using only the most similar given response to select the new response option. (We thank an anonymous reviewer for pointing this out.) It can be asked if its performance would be better if the new response option is selected by maximizing the similarity to a larger set of given responses. 

Therefore, we developed Semagent-k, a variant of the original algorithm which uses $k$-nearest neighborhood (kNN) approach for identifying the top $k$ questions in terms of highest cosine similarity of the given response and considered response options. Then the response option with the highest number of occurrences within the top $k$ question is chosen. If the two response options have the same count, the option with highest cosine similarity is taken. When there were not enough given responses available, then all the available pairs were used. The original greedy Semagent version is equivalent to Semagent-k with $k$ = 1.  

\begin{figure*}
    \centering
    \includegraphics[width=1.0\textwidth]{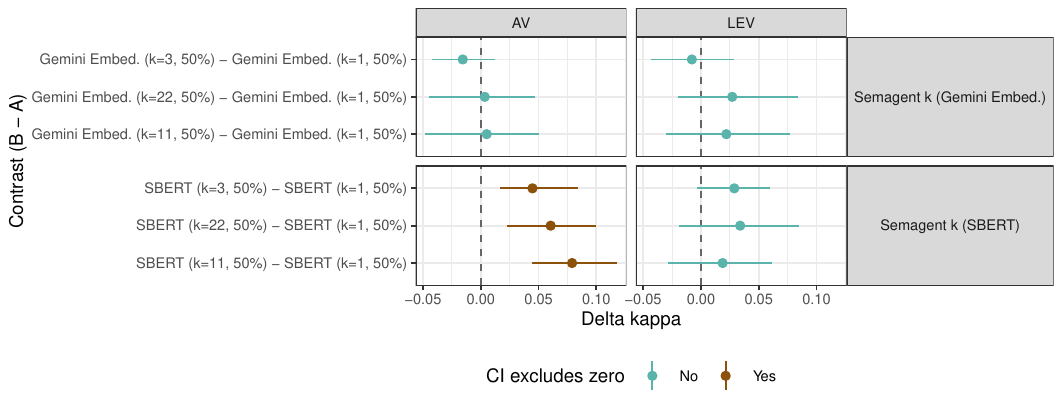}
    \caption{Bootstrapped $\Delta\kappa$ with Bonferroni corrected CIs ($\alpha=0.05$) for comparing Semagent $k$ values within AV dataset. If CI overlap zero the difference is not statistically significant. If CI are above zero, then B had statistically significantly higher $\kappa$ values than A, and vice versa.}
    \label{fig:delta_kappas_Semagent-k}
\end{figure*}

We tested $k = 1,  3, 11$ and $22$ with 50\% of known answers with the same respondents as in the main analysis (Appendix Table~\ref{tbl:kappa_Semagent_variants}). With Gemini Embeddings, the $k = 3, 11$, or $22$ did not results in a statistically significantly better $\kappa$ than $k$ = 1 (Appendix Figure~\ref{fig:delta_kappas_Semagent-k}. With SBERT, in contrast, $k$ = 1 did perform statistically significantly worse than the others.

Considering the covariance preservation, increasing $k$ resulted in worse performance (smaller RV coefficient and larger Frobenius norm) (Appendix Table~\ref{table:Semagent_variants_covariance}. Consequently, $k$ = 1 was used when deriving the semantic baselines.

\begin{table*}[t]
\caption{$\kappa$ and 95\% CIs for Semagent generated data with varying k.}
\label{tbl:kappa_Semagent_variants}
\small
\begin{tabular*}{\linewidth}{@{\extracolsep{\fill}}lllrrr}
\toprule
Dataset & Model & Variant & Kappa & Lower CI & Upper CI \\
\midrule\addlinespace[2.5pt]
AV & Gemini Embed. & k=1 & 0.413 & 0.375 & 0.449 \\
AV & Gemini Embed. & k=3 & 0.397 & 0.356 & 0.433 \\
AV & Gemini Embed. & k=11 & 0.418 & 0.377 & 0.464 \\
AV & Gemini Embed. & k=22 & 0.416 & 0.376 & 0.466 \\
AV & SBERT & k=1 & 0.336 & 0.300 & 0.371 \\
AV & SBERT & k=3 & 0.381 & 0.346 & 0.419 \\
AV & SBERT & k=11 & 0.415 & 0.379 & 0.452 \\
AV & SBERT & k=22 & 0.397 & 0.360 & 0.432 \\
LEV & Gemini Embed. & k=1 & 0.378 & 0.326 & 0.429 \\
LEV & Gemini Embed. & k=3 & 0.370 & 0.318 & 0.434 \\
LEV & Gemini Embed. & k=11 & 0.400 & 0.342 & 0.448 \\
LEV & Gemini Embed. & k=22 & 0.405 & 0.352 & 0.451 \\
LEV & SBERT & k=1 & 0.350 & 0.297 & 0.400 \\
LEV & SBERT & k=3 & 0.379 & 0.336 & 0.429 \\
LEV & SBERT & k=11 & 0.369 & 0.315 & 0.422 \\
LEV & SBERT & k=22 & 0.384 & 0.338 & 0.428 \\
\bottomrule

\end{tabular*}
\end{table*}

\begin{table*}[t]
\caption{Semagent covariance preservation with different $k$ values.}
\label{table:Semagent_variants_covariance}
\small
\begin{tabular*}{\linewidth}{@{\extracolsep{\fill}}lllrr}
\toprule
Dataset & Model & Variant & RV Coefficient & Frobenius Norm \\
\midrule\addlinespace[2.5pt]
AV & Gemini Embed. & k=1 & 0.94 & 5.14 \\
AV & Gemini Embed. & k=3 & 0.91 & 5.81 \\
AV & Gemini Embed. & k=11 & 0.88 & 9.20 \\
AV & Gemini Embed. & k=22 & 0.87 & 9.70 \\
AV & SBERT & k=1 & 0.90 & 6.38 \\
AV & SBERT & k=3 & 0.88 & 7.62 \\
AV & SBERT & k=11 & 0.85 & 11.60 \\
AV & SBERT & k=22 & 0.84 & 12.13 \\
LEV & Gemini Embed. & k=1 & 0.98 & 2.49 \\
LEV & Gemini Embed. & k=3 & 0.98 & 2.46 \\
LEV & Gemini Embed. & k=11 & 0.95 & 6.77 \\
LEV & Gemini Embed. & k=22 & 0.95 & 6.82 \\
LEV & SBERT & k=1 & 0.96 & 4.26 \\
LEV & SBERT & k=3 & 0.95 & 5.02 \\
LEV & SBERT & k=11 & 0.93 & 7.48 \\
LEV & SBERT & k=22 & 0.93 & 7.01 \\
\bottomrule

\end{tabular*}
\end{table*}

\clearpage

\section{Factor structure and item-level biases}\label{app:factor_structure}

Factor structure preservation was investigated by fitting the previously confirmed factor structure to the imputed datasets using confirmatory factor analysis (CFA). CFA fit indices are given in Appendix Table~\ref{table:cfa_indices}. All LLM imputed dataset have fit indices similar to the human datasets, expect for the cases where the model did not converge. Semagent imputed datasets did not converge. 

Analysis of item-level biases give additional insights to the covariance and factor structure preservations. We computed Vargha and Delaney's $A$ effect size, a measure of probability of superiority, by comparing LLM-generated responses with human baseline responses. Values range from 0 to 1, where $A = 0.5$ indicates identical distributions, $A < 0.5$ indicates systematically lower LLM-generated scores relative to the human baseline, and $A > 0.5$ indicates systematically higher scores. The result show that each LLM have their own systematic tendency to rate some items higher or lower regardless the share of known responses. For Semagent data, the magnitude of the bias is particularly strong for some items.

\begin{table}[t]
\caption{CFA indices for human data and imputed datasets. Values with n/a resulted in non-convergent fit.}
\label{table:cfa_indices}
\small
\begin{tabular*}{\linewidth}{@{\extracolsep{\fill}}llrrrr}
\toprule
Dataset & Source & Known answers & CFI & RMSEA & Overall MAE \\
\midrule\addlinespace[2.5pt]
AV & Human & 100 \% & 0.986 & 0.099 & 0.000 \\
AV & GPT-5.6 luna & 20 \% & n/a & n/a & n/a \\
AV & GPT-5.6 luna & 50 \% & 0.997 & 0.094 & 0.078 \\
AV & GPT-5.6 luna & 80 \% & 0.989 & 0.106 & 0.035 \\
AV & Gemini 3.7 Flash & 20 \% & n/a & n/a & n/a \\
AV & Gemini 3.7 Flash & 50 \% & 0.996 & 0.120 & 0.090 \\
AV & Gemini 3.7 Flash & 80 \% & 0.988 & 0.109 & 0.043 \\
AV & Mistral 4 Small & 20 \% & 0.999 & 0.063 & 0.079 \\
AV & Mistral 4 Small & 50 \% & 0.997 & 0.071 & 0.055 \\
AV & Mistral 4 Small & 80 \% & 0.993 & 0.080 & 0.023 \\
LEV & Human & 100 \% & 0.964 & 0.089 & 0.000 \\
LEV & GPT-5.6 luna & 20 \% & n/a & n/a & n/a \\
LEV & GPT-5.6 luna & 50 \% & 0.949 & 0.196 & 0.065 \\
LEV & GPT-5.6 luna & 80 \% & 0.943 & 0.124 & 0.035 \\
LEV & Gemini 3.7 Flash & 20 \% & n/a & n/a & n/a \\
LEV & Gemini 3.7 Flash & 50 \% & n/a & n/a & n/a \\
LEV & Gemini 3.7 Flash & 80 \% & 0.935 & 0.133 & 0.032 \\
LEV & Mistral 4 Small & 20 \% & 0.946 & 0.141 & 0.071 \\
LEV & Mistral 4 Small & 50 \% & 0.947 & 0.118 & 0.043 \\
LEV & Mistral 4 Small & 80 \% & 0.955 & 0.099 & 0.017 \\
\bottomrule

\end{tabular*}
\end{table}

\begin{figure*}
    \centering
    \includegraphics[width=1.0\textwidth]{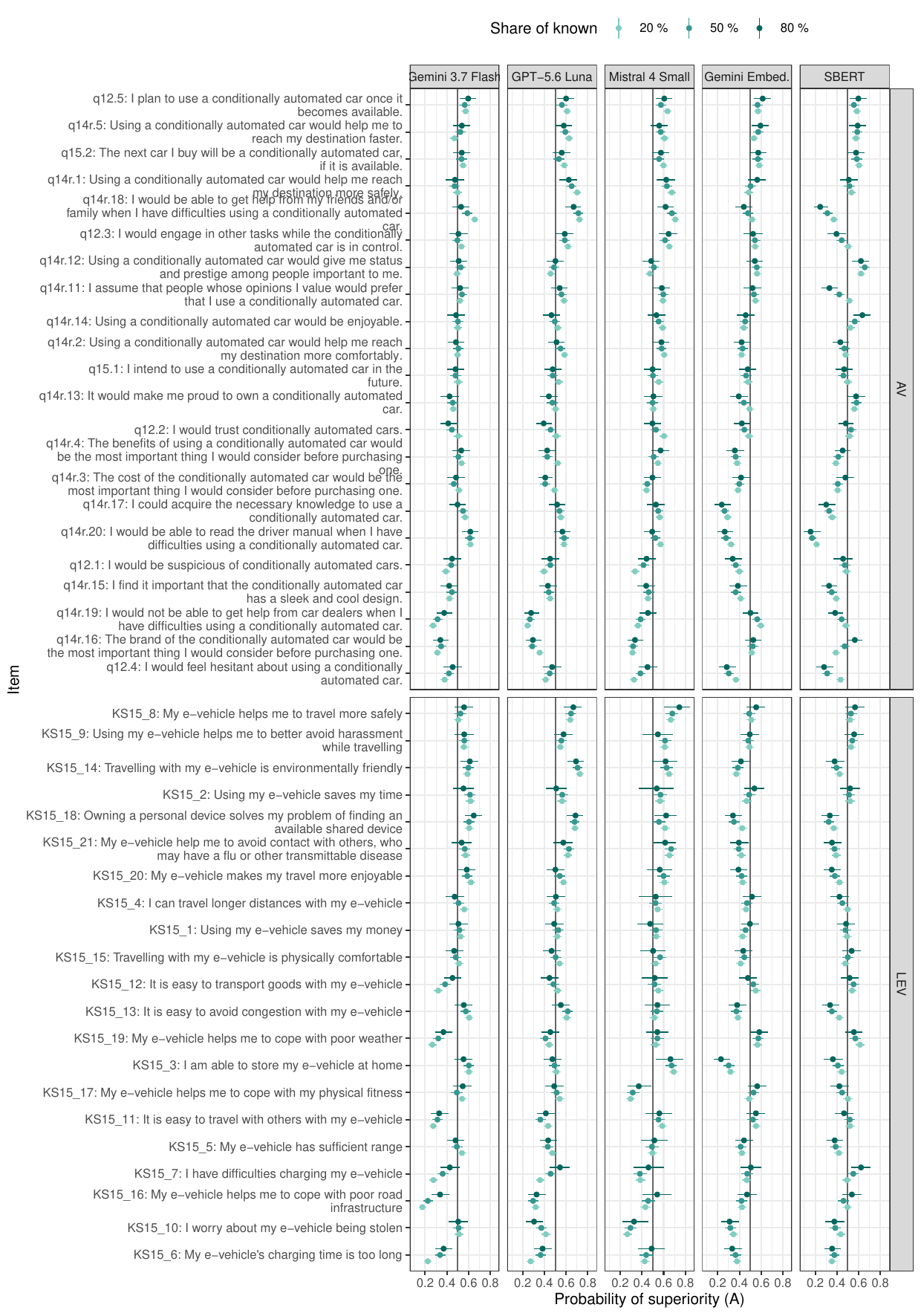}
    \caption{Probability of superiority for each item in the predicted data estimated with Vargha and Delaney's $A$ with 95\% CIs.}
    \label{fig:item_level_bias}
\end{figure*}

\end{document}